\documentclass[conference]{IEEEtran}
\IEEEoverridecommandlockouts
\pdfoutput=1
\usepackage{cite}
\usepackage[normalem]{ulem}
\usepackage{amsmath,amssymb,amsfonts}
\usepackage{algorithmic}
\usepackage{graphicx}
\usepackage{textcomp}
\usepackage{xcolor}
\usepackage{multirow} 
\def\BibTeX{{\rm B\kern-.05em{\sc i\kern-.025em b}\kern-.08em
    T\kern-.1667em\lower.7ex\hbox{E}\kern-.125emX}}
    
\begin{document}

\title{On the effectiveness of convolutional autoencoders on image-based personalized recommender systems \\
\thanks{This research has been financially supported in part by European Union ERDF funds, by the Spanish Ministerio de Econom\'ia y Competitividad (research project TIN2015-65069-C2), by the Xunta de Galicia (research projects GRC2014/035 and ED431G/01), and by the Principado de Asturias Regional Government (research project IDI-2018-000176). We would like to express our gratitude to the CESGA for the provided resources that allowed this research, and the support of NVIDIA Corporation with the donation of the Titan Xp GPUs.
}}

\author{\IEEEauthorblockN{Eva Blanco-Mallo}
\IEEEauthorblockA{\textit{Department of Computer Science} \\
\textit{Universidade da Coru\~na}\\
A Coruña, Spain \\
eva.blanco@udc.es}
\and
\IEEEauthorblockN{Beatriz Remeseiro}
\IEEEauthorblockA{\textit{Department of Computer Science} \\
\textit{Universidad de Oviedo}\\
Oviedo, Spain \\
bremeseiro@uniovi.es}
\and
\IEEEauthorblockN{Verónica Bolón-Canedo}
\IEEEauthorblockA{\textit{Department of Computer Science} \\
\textit{Universidade da Coru\~na}\\
A Coruña, Spain \\
vbolon@udc.es}
\and
\IEEEauthorblockN{Amparo Alonso-Betanzos}
\IEEEauthorblockA{\textit{Department of Computer Science} \\
\textit{Universidade da Coru\~na}\\
A Coruña, Spain \\
ciamparo@udc.es}
}

\maketitle
\bibliographystyle{IEEEtran}

\begin{abstract}
Recommender systems (RS) are increasingly present in our daily lives, especially since the advent of Big Data, which allows for storing all kinds of information about users' preferences. Personalized RS are successfully applied in platforms such as Netflix, Amazon or YouTube. However, they are missing in gastronomic platforms such as TripAdvisor, where moreover we can find millions of images tagged with users' tastes. This paper explores the potential of using those images as sources of information for modeling users' tastes and proposes an image-based classification system to obtain personalized recommendations, using a convolutional autoencoder as feature extractor. The proposed architecture will be applied to TripAdvisor data, using users' reviews that can be defined as a triad composed by a user, a restaurant, and an image of it taken by the user. Since the dataset is highly unbalanced, the use of data augmentation on the minority class is also considered in the experimentation. Results on data from three cities of different sizes (Santiago de Compostela, Barcelona and New York) demonstrate the effectiveness of using a convolutional autoencoder as feature extractor, instead of the standard deep features computed with convolutional neural networks.
\end{abstract}


\section{Introduction} \label{sec:introduction}
Digital revolution has undoubtedly changed our lifestyle. With the advent of e-commerce, social networks dedicated to sharing reviews and recommendations about different products started to become popular, mainly due to the large number of options that consumers had within their reach, as well as the distrust generated by not being able to directly see what is being purchased. Because of this, we can find millions of images tagged with the tastes of users in these social networking web services. 
This, among other factors, led to the integration of different personalized recommendation systems with great success in several online platforms.

On the other hand, due to the boom that culinary culture is experiencing in social networks in recent years, platforms such as TripAdvisor or FourSquare are becoming very popular. If we consider that TripAdvisor has approximately 460 million unique monthly visitors and more than 830 million opinions\footnote{https://tripadvisor.mediaroom.com/us}, we can have an idea of the economic impact of using these data in a personalized recommender system.

In recent years, the use of images in recommender systems (RS) is increasing \cite{fan2008justclick, xu2008personalized}, specifically through feature extraction \cite{he2016vbpr, kurt2017image}.
Some of the approaches that use images in their recommendations also employ text or metadata to compliment them \cite{chu2017hybrid}, others do not make personalize recommendations \cite{amis2017}, and those that use personalization, do not consider images \cite{zhang2017novel, zhang2018personalized}. To the best of our knowledge, our paper is the first one that explores the use of images in a personalized RS for restaurants, taking advantage from convolutional autoencoders for feature extraction.

Therefore, the goal of this paper is to study how well an RS works when using convolutional autoencoders to model users and items through their images. In this context, we propose a model that uses the images uploaded by users of a gastronomic platform in order to build a personalized recommendation. For doing that, we need to find the feature vector that best defines a given image, trying to improve the performance of a personalized restaurant RS. Convolutional autoencoders are commonly used in various image-related tasks, such as compression \cite{cheng2019energy}, reconstruction \cite{chaitanya2017interactive}, noise reduction \cite{xie2012image}, and feature extraction \cite{hong2015multimodal}. However, up to our knowledge, their use has not been explored in depth in the context of image-based RS. Different than existing approaches, the main contribution of this work is to use a convolutional autoencoder as feature extractor for the images that feed a personalized RS. We will demonstrate that our approach is sound because (1) it makes use of the context of the problem, (2) it works better than standard approaches that use a pre-trained convolutional neural network (CNN), and (3) it is less computationally expensive that integrating and fine-tuning a CNN.

The remainder of this paper is organized as follows. Section \ref{related_work} includes a review of the latest studies related to restaurant recommendations, as well as the use of images in RS. Section \ref{methodology} presents the architecture and implementation of the proposed model. In Section \ref{experimental_results}, we describe the three datasets used, the different experiments carried out, and the performance achieved. Finally, in Section \ref{conclusion} the most relevant aspects that we could extract from experimentation are presented, as well as various lines of interesting future research.

\section{Related Work} \label{related_work}

As mentioned in Section \ref{sec:introduction}, there are several attempts aiming to build an RS based on available TripAdvisor reviews. Zhang et al. \cite{zhang2017novel, zhang2018personalized} used TripAdvisor data, but without considering the visual information at all. Amis \cite{amis2017} used TripAdvisor images to select the photos to be displayed in search results employing a convolutional network, but not in a personalized way. Regarding restaurant recommendations, Chu et al. \cite{chu2017hybrid} investigated the visual effects of images using blog photographs along with the accompanying text. Feature extraction was carried out using a CNN, but with a support vector machine to previously classify images into different categories. Therefore, we have not found other related studies in the available literature with which to compare our results. 

Regarding the use of visual information on personalized RS, one of the first attempts was proposed by He et al. \cite{he2016vbpr}, using a CNN to extract the deep features of product images, which are then processed in an RS based on matrix factorization. Notice that the main difference concerning our proposal is that we intend to use images in order to define not only the user profile but also the item to recommend. Fant et al. \cite{fan2008justclick} developed an imaged-based RS to allow personalized recommendations via exploratory search, from large-scale collections of manually-annotated Flickr images. These collections were first organized at a semantic level, and then series of points of interest and their scale-invariant feature transform (SIFT) features were used to characterize the local visual properties of the images. Kurt et al. \cite{kurt2017image} proposed an image-based RS using the bag of words technique and feature descriptors such as SIFT, SURF (speeded-up robust features) and LBP (local binary patterns) \cite{kashifdescrptors} to characterize shoes images of the users. Finally, Xu et al. \cite{xu2008personalized} tried to predict user attention for images, dividing them into segments and using content similarity.

\section{Methodology} \label{methodology}
This work proposes a method to provide gastronomic recommendations based on users’ tastes that can be inferred from the images posted by them. In general terms, the problem at hand can be defined as a classification task in which we have some triads with either one of two labels:

\begin{equation}
(u, r, i) \leadsto 0|1,
\end{equation}
where $u$ is a user that visited the restaurant $r$ and took the image $i$. Regarding the two labels, 0 means that the user $u$ does not like the restaurant $r$, whilst 1 is the opposite.

\begin{figure}
\centerline{\includegraphics[height=3.6in]{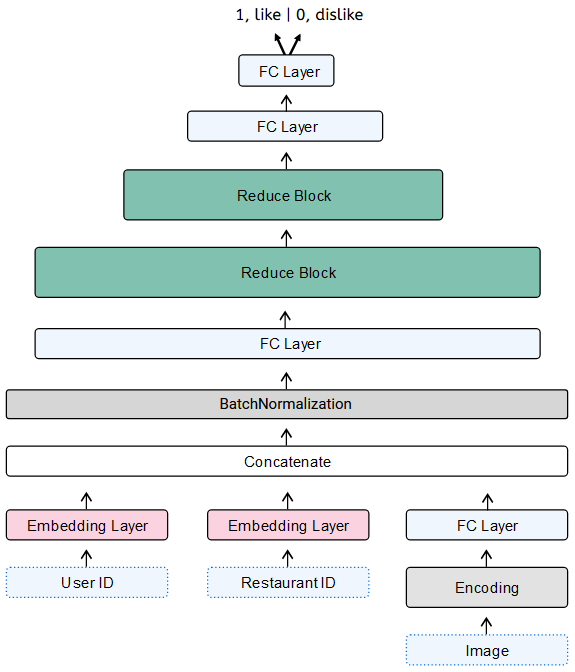}}
\caption{Architecture of the proposed method to provide image-based personalized recommendations.} \label{fig:model_arq}
\end{figure}

Aiming at solving this binary classification task, we propose the model depicted in Figure \ref{fig:model_arq}: a network that learns on triads of users, restaurants and photos $(u, r, i)$ to provide an image-based personalized recommendation. The codification of each input triad is defined as follows:

\begin{itemize}
    \item Users and restaurants are represented by a one-hot codification.
    \item Photos are represented using a convolutional encoder, next described in Section \ref{sec:autoencoder}.
\end{itemize}

The three codifications are then mapped to their corresponding embedding representations, each one composed  of 512 features, by using embedding layers for users and restaurants, and a fully connected (FC) layer for images. These three embedding codes are then concatenated, obtaining a 1536-dimensional vector. Given that the features of each individual input $(u, r, i)$ are in different scales, a batch normalization \cite{ioffe2015batch} layer is used to normalize this vector, thus speeding up the learning process. This layer is followed by a FC layer that maps the input vector into a 1024-dimensional vector.

\begin{figure}
\centerline{\includegraphics[height=3.8in]{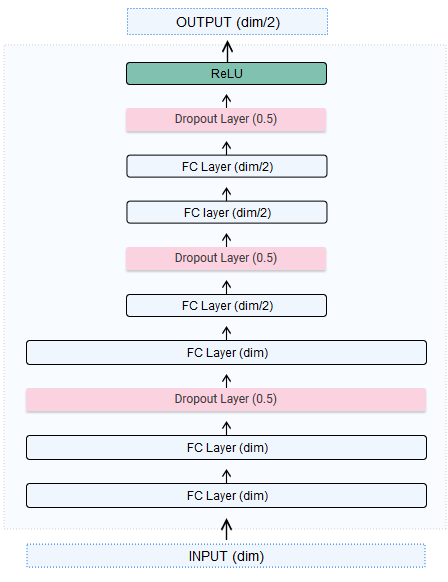}}
\caption{Reduce block structure} \label{fig:rb_arq}
\end{figure}

The vector obtained includes an individual representation of each element in the triad, $(u, r, i)$, and so a subsequent processing is necessary to obtain a joint representation. This step is carried out by means of two reduction blocks, whose structure is illustrated in Figure \ref{fig:rb_arq}. As can be observed, a reduce block maps an input vector into an output vector of half size, and is composed of a sequence layers that include FC, dropout \cite{srivastava2014dropout} with a probability $p=0.5$, and rectified linear unit (ReLU) \cite{nair2010rectified} as the activation function.

After the two reduction blocks in Figure \ref{fig:model_arq}, there is a FC layer that reduces the size of the vector by half; and finally, there is another FC layer with a sigmoid activation function that generates a probability output in the range $[0, 1]$, where 0 means \textit{dislike}, and 1 \textit{like}.

\subsection{Image encoding} \label{sec:autoencoder}

As mentioned above, the proposed network needs a vectorial representation of the input image to combine the visual information with the user and restaurant codifications. For this purpose, we propose to use a convolutional encoder.

An autoencoder \cite{goodfellow2016deep} is an unsupervised machine learning algorithm that takes the input data and aims to reconstruct them back using a lower dimensional representation. It has a symmetrical structure formed by two main components: the encoder, and the decoder. The encoder transforms the input $x$ into a representation $h$, also known as \textit{code}, using a deterministic function of the type 

\begin{equation}
h = f_{\theta}(x) = \sigma (W x + b),
\end{equation}
with parameters $\theta = \{W,b\}$, where $W$ is the weight matrix, $b$ is the bias vector, and $\sigma$ is an element wise activation function. 
The decoder uses the code $h$ to generate $r$, a
reconstruction of the input $x$, using a reverse mapping of $f$: 

\begin{equation}
r = g_{\theta'}(h) = \sigma' (W' h + b'),    
\end{equation}
with parameters $\theta' = \{W',b'\}$, where $W'$ is the weight matrix, $b'$ is the bias vector, and $\sigma'$ is an element wise activation function.

It is worth noting that the autoencoder is trained to minimize the difference between input $x$ and its reconstruction $r$. Based on the premise that the reconstruction generated by the autoencoder is good enough, we can use the code generated by the encoder as a feature vector that represents the input.

\begin{figure}
\centerline{\includegraphics[height=1.8in]{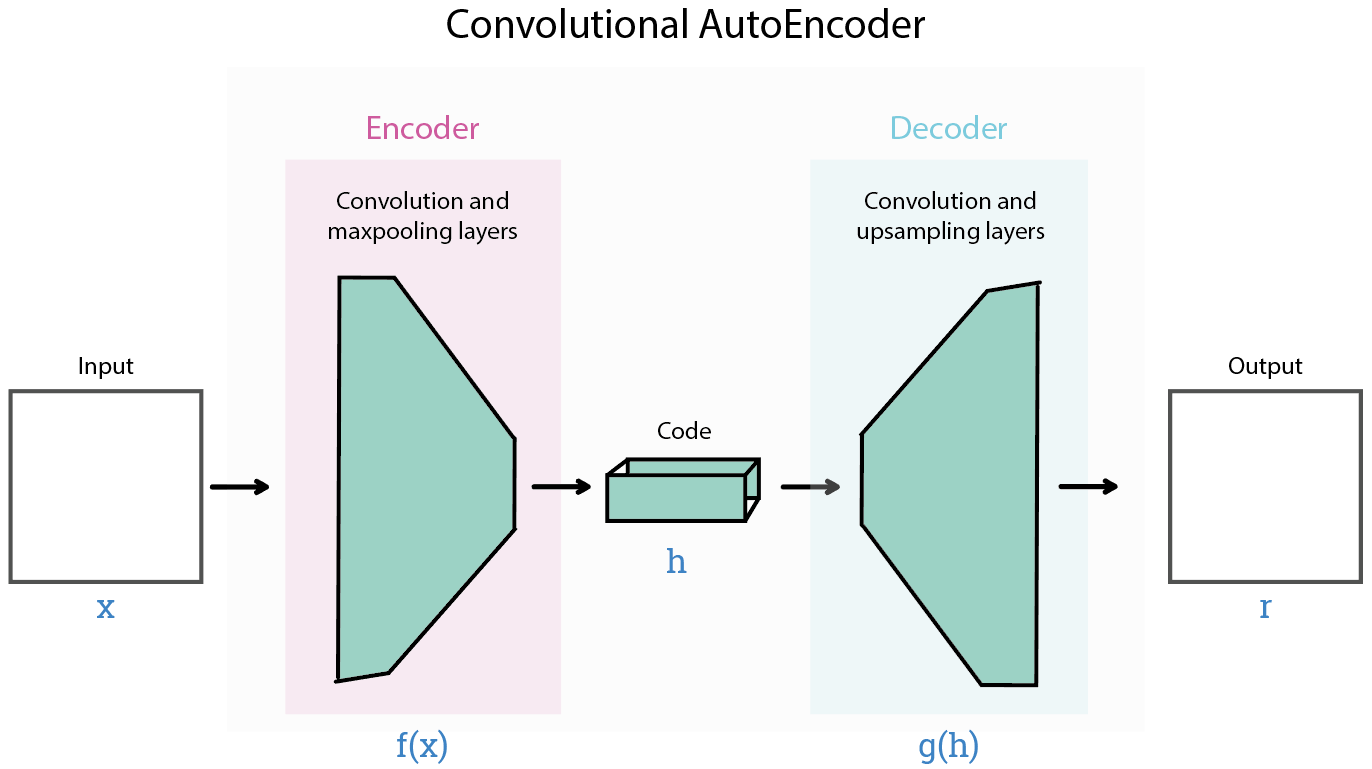}}
\caption{General architecture of a convolutional autoencoder (CAE).} \label{fig:auto_arch}
\end{figure}

When images are used as input data, we have to talk about convolutional autoencoders (CAE) \cite{masci2011stacked}. Broadly speaking, they have the same conceptual structure than any other autoencoder. Therefore, when a CAE is trained, its encoder part can be used as a highly effective feature extractor \cite{goodfellow2016deep}. The main difference is that the encoder uses convolutional and pooling layers to extract features and reduce the size of the input volume (image), respectively, to finally produce a lower-dimensional representation (code); and next, the decoder takes the code and process it to obtain the final reconstruction by means of convolutional and upsampling layers. Figure \ref{fig:auto_arch} illustrates the general architecture of a CAE. 

Table \ref{table_arch_auto} details the architecture of the CAE used in this research, based on the one proposed by Chollet \cite{arqAutoKeras}. The core part of the CAE is a building block that consists of a sequence of three layers: convolutional, batch normalization, and ReLU. Regarding the whole architecture, it is composed of a downsampling path (encoder) that includes a sequence of convolutional and maxpooling layers, and an upsampling path (decoder) that includes a sequence of convolutional and upsampling layers. The code generated by the CAE corresponds to the feature maps of the bottle neck, i.e., the intermediate representation obtained between both paths. Once the CAE is trained, its encoder part can be used as a feature extractor. More specifically, the encoder is followed by a flatten layer that converts the code generated (feature maps) into a feature vector, which represents the image encoding used in our model (see Figure \ref{fig:model_arq}).

\begin{table}
\renewcommand\arraystretch{1.3}
\begin{center}
{\caption{Architecture of the CAE used in the proposed model. }\label{table_arch_auto}}
\begin{tabular}{l|l}
\hline
\textbf{Encoder} (downsampling path) & \textbf{Decoder} (upsampling path) \\\hline
\\[-6pt]
\quad Input, $m=3$ & \quad Building Block, $m=16$\\
\quad Building Block, $m=64$ & \quad UpSampling Layer, $m=16$ \\
\quad MaxPooling Layer, $m=64$  & \quad Building Block, $m=32$ \\
\quad Building Block, $m=32$ & \quad UpSampling Layer, $m=32$ \\
\quad MaxPooling Layer, $m=32$ & \quad Building Block, $m=64$ \\
\quad Building Block, $m=16$ & \quad UpSampling Layer, $m=64$ \\
\quad Building Block, $m=3$ & \quad Convolutional Layer 3x3, $m=3$ \\
\quad MaxPooling Layer, $m=3$ & \quad Batch Normalization Layer, $m=3$ \\
\quad & \quad Activation Layer (Sigmoid), $m=3$\\[+4pt] \hline 
\multicolumn{2}{l}{Building block: $3\times3$ convolution + batch normalization + ReLU} \\
\multicolumn{2}{l}{MaxPooling layer: $dim/2$. UpSampling layer: $dim \times 2$} \\
\multicolumn{2}{l}{$dim$: spatial dimensions of the input volume} \\
\multicolumn{2}{l}{$m$: number of feature maps at the end the block or layer} \\
\end{tabular}
\end{center}
\end{table}

\section{Experimental results}
\label{experimental_results}

This section first describes the datasets used to evaluate the performance of our proposed method. Next, we detail the experimentation carried out, which includes two alternative approaches to compute the feature vector of input images. Finally, the results obtained are presented and analyzed in depth, including an ablation study.

\subsection{Dataset} \label{subsec_dataset}

The data used in this work were collected in 2018 and 2019 from the TripAdvisor reviews published by users about restaurants in cities of different sizes: 1) Santiago de Compostela (Spain), a small city located in the Atlantic coast, with a population of 95800 inhabitants; 2) Barcelona (Spain), one of the largest and most touristic cities in the Mediterranean coast, with a population of 1.7 millions of inhabitants; and 3) New York (USA), a very large and popular city of approximately 8.4 millions of inhabitants, located in the East Coast. Given that we need to store the relationships between users and restaurants, large sparse matrices are generated. As the number of individuals increases, the problem becomes more latent. Considering that the number of users and items at a city level is representative enough to make a good recommendation, three different datasets are considered, one per city.

Table \ref{table_info_data} depicts the magnitude of the three datasets, including the number of users, restaurants and reviews made by users. Notice that this research is focused on images, thus only the reviews with images are considered for experimentation. Therefore, the number of reviews are: 7003 in Santiago, 66904 in Barcelona, and 111415 in New York. As each review can include one or more images, the number of photos available for experimentation is as follows: 16168 in Santiago, 153707 in Barcelona, and 234689 in New York.

\begin{table}
\renewcommand\arraystretch{1.3}
\begin{center}
{\caption{Summary of figures of the three datasets used in the experimentation.}\label{table_info_data}}
\begin{tabular}{lllrrr}
\hline
\multicolumn{3}{l}{}&\textbf{Santiago}&\textbf{Barcelona}&\textbf{New York}\\
\hline
\multicolumn{3}{l}{Total Users}&25456&184307&61019 \\
\multicolumn{3}{l}{Total Restaurants}&513&7602&7588 \\
\multicolumn{3}{l}{Total Reviews}&43627&466964&1008761 \\
\hline
\multicolumn{3}{l}{Average reviews* per user}&1.4&1.99&1.82 \\
\multicolumn{3}{l}{No. users with one review* }&4035&23978&43095 \\ \hline
\multirow{2}*{No. reviews} & \multicolumn{2}{l}{without images}  & 36624 & 400060 & 897346\\
& \multicolumn{2}{l}{with images} & 7003 & 66904 & 111415 \\ \hline
\multirow{2}*{No. images} & \multicolumn{2}{l}{positives} & 13915 & 128605 & 207253 \\
& \multicolumn{2}{l}{negatives} & 2253 & 25102 & 27436 \\
& \multicolumn{2}{l}{\textbf{Total Images}} & \textbf{16168} & \textbf{153707} & \textbf{234689} \\ \hline 
\multicolumn{6}{l}{*Only reviews with images are considered}  \\
\end{tabular}
\end{center}
\end{table}

These datasets present some peculiarities that may hinder the recommendation task. According to the figures presented in Table \ref{table_info_data}, there is a high imbalance between the positive class (1, \textit{like}), and the negative class (0,\textit{ dislike}). In particular, the ratios between positive and negative classes are approximately 6:1 for Santiago, 5:1 for Barcelona, and 7:1 for New York. Consequently, it will be more difficult for any system to learn how to distinguish cases corresponding to negative samples. 

Regarding the data available on TripAdvisor, it is worth noting that each review includes the identifier of the user who published it, the identifier of the restaurants reviewed, a set of images (optional), and a score in terms of stars (from one to five). As detailed above, only reviews with images are used in this research and thus, each review considered contains at least one image. Regarding the scores, those reviews with from one to three stars are labeled as 0 (\textit{dislike}), whilst four or five stars are labeled as 1 (\textit{like}). Note that these labels are the output of the binary classification system proposed in Section \ref{methodology}.

\begin{figure}[ht]
\centerline{\includegraphics[height=2.7in]{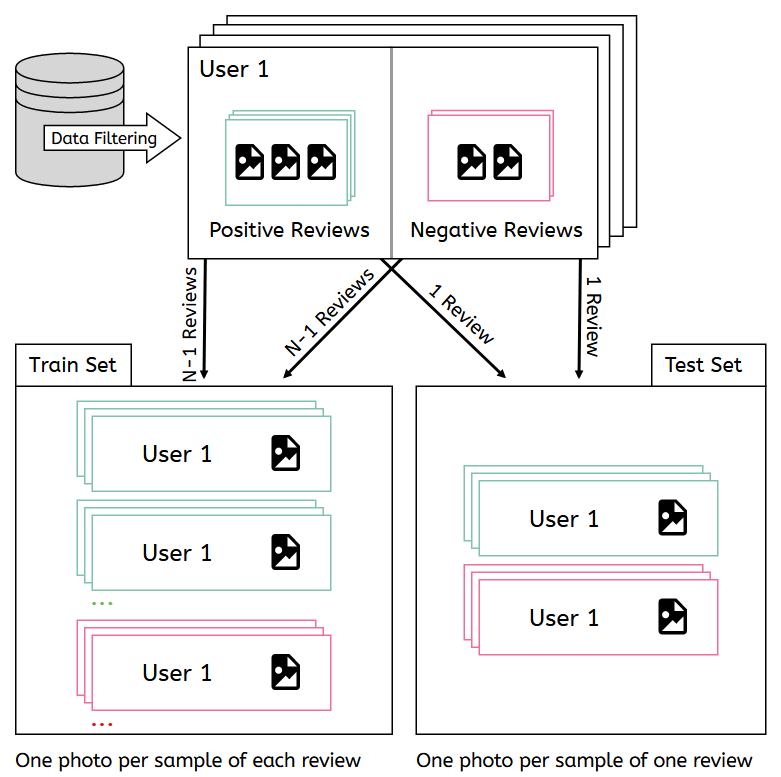}}
\caption{Splitting procedure to create the train and test sets.} \label{fig:get_patitions}
\end{figure}

For experimentation purposes, the datasets were split in train and test sets. Figure \ref{fig:get_patitions} illustrates the procedure carried out to create these partitions. Note that it is a user-dependent problem and, thus, some restrictions must be considered. The complete procedure is detailed as follows. For each user:

\begin{enumerate}
    \item If there are several reviews that belong to the same pair $(u,r)$, all the corresponding images are fed to the train set in order to avoid opposite ratings.
    \item The remaining reviews of the user $u$ are divided into two groups based on their ratings, positive or negative. For each group, all the images of one review are assigned to the test set and the images belonging to the rest of the reviews to the train set. Note that if a user has a single review, then it is assigned to the train set. In this manner, it is guaranteed that all the users evaluated in the test set have been considered in the learning process with the train set.
    \item If there is any review in the test set that belongs to a restaurant that is not included in the train set, then all its images are moved to the train set. Again, the idea is to guarantee that all the restaurants evaluated in the test set are also in the train set.
\end{enumerate}

Due to experimentation requirements, the initial train set was split again following the same procedure, thus obtaining a new train set and a validation set. From now on, we will refer to this new train set as the train set. Table \ref{table_datasets} shows the number of images per partition for the three datasets, including the ratios between positive and negative samples.

\begin{table}
\begin{center}
\renewcommand\arraystretch{1.3}
{\caption{Number of samples (images) per city for the three partitions: train, validation, and test.}\label{table_datasets}}
\begin{tabular}{clrrr}
\hline
&&train&val&test\\ \hline 
\multirow{5}{0.5cm}{\rotatebox{90}{\textbf{Santiago}}} & \quad Positive samples&11443&675&1797\\
&\quad Negative samples&1913&87&253\\ 
&\quad Total samples&13356&762&2050\\ [1pt]
&\quad Ratio &5.98:1&7.76:1&7.1:1 \\ [2pt]
\hline 
\multirow{5}{0.5cm}{\rotatebox{90}{\textbf{Barcelona}}} & \quad Positive samples&101275&9749&17581\\
&\quad Negative samples&19805&1842&3455\\ 
&\quad Total samples&121080&11591&21036\\ [1pt]
&\quad Ratio &5.11:1&5.29:1&5.09:1 \\ [2pt]
\hline 
\multirow{5}{0.5cm}{\rotatebox{90}{\textbf{New York}}} &\quad Positive samples&157323&16560&33470\\
&\quad Negative samples&22693&1397&3346\\
&\quad Total samples&180016&17957&36816\\ [1pt]
&\quad Ratio &6.93:1&11.85:1&10:1 \\ [2pt]
\hline
\end{tabular}
\end{center}
\end{table}

As the datasets are highly unbalanced, a strategy must be applied to reduce its impact on the model performance. Data augmentation \cite{perez2017effectiveness} consists in increasing the number of samples (images) of the train set, without collecting new data. Particularly, image data augmentation involves expanding the size of the train set by creating modified versions of the original images. The objective of this technique is not only to increase the amount of data available, but also their variability, thus improving the robustness of the learning models. 

Data augmentation can be applied to all the samples in the train set but, following an over-sampling perspective \cite{rahman2013addressing}, we only over-sampled the minority class by applying four different transformations: rotation (-30º), flipping ($x$ axis), re-scaling (1.25) and translation ([5,5]). As a result, the imbalance problem is alleviated and the ratios between the positive and the negative classes are very close to 1:1 (1.2:1 for Santiago de Compostela, 1.02:1 for Barcelona, and 1.39:1 for New York). Notice that all the experiments that entail training the proposed model use the augmented train set.

\subsection{Experiments} \label{sec:experiments}

This section presents the details of the experimentation carried out to evaluate the proposed model, showing how an RS that models users' preferences by means of their images works in terms of prediction. It also describes the two baseline methods considered to provide a comparative study in order to contrast the effectiveness of using a convolutional encoder as feature extractor. Both of them are based on CNNs, the most commonly used approach in the related work described in Section \ref{related_work}.

CNNs are considered a benchmark in the supervised classification of images since Krizhevsky et al. won the \textit{ImageNet Large Scale Visual Recognition Challenge} (ILSVRC) with the architecture AlexNet \cite{krizhevsky2012imagenet}; and they have been successfully applied to different computer vision tasks, such as object detection \cite{ren2015faster,aguilar2018grab} 
or image segmentation \cite{long2015fully,chen2017deeplab}. 
In terms of image classification, a CNN is composed of, among others, convolutional layers that extract features followed by fully connected layers that perform the final classification. Notice that, once the CNN is trained, the convolutional base of the network can be used as a feature extractor.

The baseline methods are used to measure the impact of using a CAE in the image encoding step (see Figure \ref{fig:model_arq}). Both methods use the convolutional base of the ResNet50 \cite{he2016deep}, with weights pre-trained on ImageNet\footnote{https://keras.io/applications/\#resnet}. In the first scheme, the pre-trained ResNet50 is used as feature extractor to generate the codification of the images. That is, the image encoding of our model is performed by the convolutional base of the ResNet50, instead of using the CAE. In the second one, ResNet50\_FT, the convolutional base of the ResNet50 is integrated in our model to get the image encoding. In this manner, its weights are fine-tuned for the problem at hand. 

In order to evaluate the performance of the proposed model, either using the feature vector generated by the CAE or the CNN, a complete set of performance measures was used:
\begin{itemize}
    \item Sensitivity, also known as recall: probability that the system correctly classifies a sample of the positive class.
    \item Specificity: probability that the system correctly classifies a sample of the negative class.
    \item Precision: ratio of positive samples correctly classified among the total of samples classified as positive.
    \item F1-Score: the harmonic mean of precision and recall. 
\end{itemize}

Precision and recall are two of the most popular metrics in RS. Precision does not work well for unbalanced datasets, as it is our case, and then F1-score is commonly used in these scenarios. The main problem with the F1-score is that it gives more importance to the positive class, and thus problems with the negative class may go unnoticed. In a recommendation system, not only it is important to recommend those items that the user likes, but also not to recommend those items that the user does not like. Therefore, we look for a system with high values for both sensitivity and specificity. For this purpose, we propose to use the balanced score (B-score), a variation of the F1-score that calculates the harmonic mean of sensitivity and specificity: 
\begin{equation}
\text{B-score} = 2 * \frac{\text{sensitivity} * \text{specificity}}{\text{sensitivity} + \text{specificity}}
\label{b_score}
\end{equation}

If we rely on metrics such as precision or F1-score to select the best model, we could choose the one that best detects the items that the user likes, even if the performance is very low when classifying the items that he/she does not like (e.g., 0.99 sensitivity and 0.40 specificity: F1-score = 0.94, B-score = 0.56), rather than one with more balanced and desirable predictions (e.g., 0.85\% sensitivity and 0.70\% specificity: F1-score = 0.89, B-score = 0.76). Using the B-score, the model chosen will be the one with the best balance between sensitivity and specificity. Although we focus on the B-score value throughout the experiments, and sensitivity and specificity are shown for a more complete quick view of the results, we also keep precision and F1-score since they are standard metrics. However, it is important to emphasize that these metrics do not correctly represent the behavior we are looking for.

All the experiments were performed on a computer equipped with a GeForce Titan XP 12GB GPU from NVIDIA, an Intel Core i7-4790 CPU @ 3.60GHz x 8, and 16 GiB memory. The implementation of the model and baseline methods is in Keras\footnote{https://keras.io/}, with Tensorflow as backend, and the code will be publicly available after paper acceptance. The models' weights were initialized using the HeUniform initialization \cite{he2015delving}, except for the weights of the ResNet50, as previously mentioned; and the Adam algorithm \cite{kingma2015adam} was used as optimizer. The training process was carried out by setting a batch size of 32, and the outputs were monitored by using the B-score (see Eq. \ref{b_score}), with a patience of 12 epochs and a maximum of 100 epochs. Note that the original images were resized to $224 \times 224 \times 3$ to meet the input size requirements of the ResNet50. Although CAEs do not impose any input size restriction, we considered the same input size when applying the architecture of Table \ref{table_arch_auto}.

We performed a grid search to determine the best hyper-parameters of our system, using the validation scheme described in Section \ref{subsec_dataset}, monitoring the learning process with a patience of 6, and focusing on the results obtained in the train and validation sets. After trying different options for the learning rate (0.001, 0.0001), and dimension of user/restaurant embeddings (128, 256, 512), the best ones were 0.001 and 512, respectively. Once these hyper-parameters were established, the models were trained with the train set and next evaluated with the test set. Moreover, an ablation study was carried out to verify the effect of reducing the depth of the model by eliminating one of the two reduce blocks.

\subsection{Results} \label{subsec_results}

In order to train and evaluate the proposed model, the autoencoder must be trained first. For its training, the original train set (i.e., the train set without data augmentation) is used, although later on, the augmented train set is used for model training. As detailed in Section \ref{subsec_dataset}, input images were resized to $224 \times 224 \times 3$. Taking into account the input dimensions and the architecture proposed in Table \ref{table_arch_auto}, the encoder part generates three feature maps of size $28\times28$, thus resulting in a 2352-dimensional feature vector used as image encoding.

\begin{figure}
\centerline{\includegraphics[width=\columnwidth]{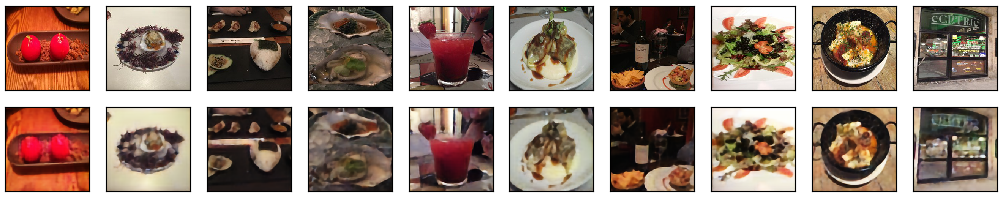}}
\caption{Representative samples of the Barcelona test set: input images (first row), and their corresponding outputs (second row) generated by the trained autoencoder} \label{fig:auto_test}
\end{figure}

Regarding the learning process, the autoencoder was trained with a batch size of 32, a patience of 6 and a maximum of 100 epochs, using the loss function to monitor the results. Figure \ref{fig:auto_test} illustrates some examples of input images and their respective outputs generated by the trained autoencoder.

In order to demonstrate the adequacy of our model based on autoencoders, we provide a comparison with two baseline methods based on CNNs (see Section \ref{sec:experiments}). Table \ref{table_comp} shows the comparative results with the three approaches considered: the model proposed in this research, which uses an autoencoder as feature extractor; and two alternative methods that replace the autoencoder with CNNS (pre-trained ResNet50 as feature extractor, and integrated and fine-tuned ResNet50). The three approaches were tested on the three different datasets (Santiago de Compostela, Barcelona, and New York). As can be observed, our proposal obtains the best trade-off between sensitivity and specificity for the three datasets, i.e., it provides the best performance in terms of B-score. In contrast, the less competitive approach is the one that uses the pre-trained ImageNet without fine-tuning (ResNet50), the only one that was trained \textit{out of context}. In particular, the ResNet50 approach shows problems to correctly classify the minority class, providing very low specificity values. When applying fine-tuning (ResNet50\_FT), the results improved, since this approach includes a fine-tuning of the ResNet50 weights. In this sense, the convolutional layers are able to adapt to the context using the information associated with the images and become more specialized, being able to discard irrelevant information in the problem at hand, which would be inevitably used when applying transfer learning.

Analyzing the differences between the three datasets, it can be seen that the autoencoder works better that the other two approaches, in terms of specificity, when more data is available. In fact, it provides the highest specificity values in Barcelona and New York, but not in Santiago. Nevertheless, the autoencoder approach obtained the highest B-Score, balancing both sensitivity and specificity regardless the dataset considered.

Taking into account that the strongest point of our approach is the use of images as a source for modeling users' preferences, the convolutional autoencoder is crucial in this process. Compared to ResNet50, it is able to improve the classification of the minority class without incurring in significant losses when detecting the majority class. Furthermore, there is an important reduction in the runtime required to train the autoencoder compared to adjusting the weights of the filters during the training of the ResNet50\_FT, which provides the second best performance, making our proposal the most affordable one in terms of training time.

\begin{table}
\begin{center}
\renewcommand\arraystretch{1.3}
{\caption{Comparison of results using three different techniques for the image encoding step of the proposed model.}\label{table_comp}}
\begin{tabular}{clccc}
\hline
&&Autoencoder&ResNet50&ResNet50\_FT\\
\cline{1-5}
\multirow{5}{0.5cm}{\rotatebox{90}{\textbf{Santiago}}} & \quad Sensitivity&0.7629&0.8453&0.7318 \\
&\quad Specificity&0.7905&0.6047&0.8181 \\
&\quad Precision&0.9628&0.9382&0.9662 \\
&\quad F1-Score&0.8513&0.8893&0.9662 \\
&\quad B-Score&0.7765&0.7050&0.7725 \\
\hline
\multirow{5}{0.5cm}{\rotatebox{90}{\textbf{Barcelona}}} & \quad Sensitivity&0.6175&0.8546&0.6738 \\
&\quad Specificity&0.8006&0.4480&0.6176 \\
&\quad Precision&0.9403&0.8874&0.8997 \\
&\quad F1-Score&0.7454&0.8706&0.7705 \\
&\quad B-Score&0.6972&0.5878&0.6445 \\
\hline
\multirow{5}{0.5cm}{\rotatebox{90}{\textbf{Ney York}}} & \quad Sensitivity&0.7454&0.7261&0.7933 \\
&\quad Specificity&0.7594&0.7071&0.6724 \\
&\quad Precision&0.9687&0.9612&0.9603 \\
&\quad F1-Score&0.8425&0.8273&0.8676 \\
&\quad B-Score&0.7523&0.7165&0.7279 \\
\hline
\\ [-6pt]
\multicolumn{5}{l}{ResNet50\_FT: parameter fine-tuning} \\
\end{tabular}
\end{center}
\end{table}

\subsection{Ablation study}
\label{subsec_ablation}

An ablation study was carried out trying to assess the influence of diminishing the number of reduce blocks in the proposed model (see Figure \ref{fig:model_arq}). Table \ref{tab:astudy_rbs} shows the performance of the proposed model based on its depth, checking the effects of eliminating a reduce block from its structure. Using two reduce blocks exerts a great influence on the detection of cases belonging to the minority class, but at the cost of losing sensitivity. In the case of Santiago and Barcelona, increasing the depth of the model decreases the sensitivity (in 0.1435 and 0.0763, respectively), but highly improves the specificity (in 0.2715 and 0.0988, respectively). It is worth mentioning that when dealing with the largest dataset (New York), both sensitivity and specificity were improved when using two reduce blocks. In general, looking at the B-score metric, which balances sensitivity and specificity, the architecture with two reduce blocks is the best option for the three datasets.

\begin{table}
\begin{center}
\renewcommand\arraystretch{1.3}
{\caption{Ablation study that measures the effects of reducing the depth of the model, using one or two reduce blocks in its architecture.}\label{tab:astudy_rbs}}
\begin{tabular}{lrrrrrr}
\hline
\rule{0pt}{6pt}
\quad &\multicolumn{2}{c}{\textbf{Santiago}}&\multicolumn{2}{c}{\textbf{Barcelona}}&\multicolumn{2}{c}{\textbf{New York}} \\
\cline{2-7}
\quad &1RB&2RB&1RB&2RB&1RB&2RB \\
\rule{0pt}{6pt}
\\[-10pt]
Sens.&0.8392&0.7629&0.7610&0.6175&0.7378&0.7454 \\
Spec.&0.6917&0.7905&0.5291&0.8006&0.7304&0.7594 \\
Precision&0.9508&0.9628&0.8916&0.9403&0.9648&0.9687 \\
F1-Score&0.8915&0.8513&0.8212&0.7454&0.8362&0.8425 \\
B-Score&0.7583&0.7765&0.6242&0.6972&0.7341&0.7523 \\
\\ [-6pt]
\hline
\\ [-6pt]
\multicolumn{7}{l}{ 1RB: one reduce block. 2RB: two reduce blocks.}
\end{tabular}
\end{center}
\end{table}

\section{Conclusion} \label{conclusion}

Despite the great success that RS are showing in recent years, there is no approach, in the literature or commercially, that explores in depth the use of images in the context of personalized restaurant recommendations. In this work, we explored the potential of modeling both users and restaurants using images as single source. Specifically, we presented an image-based RS that makes use of TripAdvisor data to predict users' tastes, with the particularity that it employs a convolutional autoencoder as feature extractor. Due to the nature of the problem, with typically more positive than negative reviews, the datasets used are highly unbalanced. For this reason, we proposed to use data augmentation on the minority class. We also found out that metrics such as F1-score are not appropriate for our problem, since it does not take into account both the sensitivity and specificity. For this reason, we have used the B-score that balances both. This metric is especially recommended in this case, since we have a binary classification system with the positive class as the majority one, and the misclassification of the negative class must be penalized without passing through a high success rate in the positive class.

In the three cities considered, we achieve a B-score $\approx 70$\% in the worst case, thus showing the potential of using images in the context of personalized recommendations. Our approach, which uses an autoencoder as feature extractor, was compared with the standard approach of using CNNs for image encoding. In particular, we considered the popular ResNet50, pre-trained on ImageNet with and without parameter fine-tuning. The experimental results demonstrated the effectiveness of using the autoencoder as feature extractor, since it allowed to improve the detection of the minority class, and thus obtaining the highest B-score values for the three datasets. As expected, the worse option was to use the ResNet50 without fine-tuning because, although it has been trained with a much larger database (ImageNet), the purpose of the training was to solve a image classification task that is not related to the problem at hand; thus, it did not take advantage of the context that involves users and restaurants. Therefore, we can then conclude that the context is very important when learning to detect this type of features, having to choose quality over quantity, if necessary. 

Regarding the fine-tuned CNN, it improves the results obtained with the pre-trained CNN since, in the first case, the parameters are fine-tuned for the problem at hand. However, if we compare the fine-tuned CNN with the autoencoder, it can be seen that the former cannot achieve the competitive results of the latter. This fact seems to indicate that the feature extraction does not depend on the user, although the subsequent recommendation does, i.e., the features are intrinsic to the restaurant, not to the user. That is, taking into account the subjectivity that characterizes tastes, there are images of the same restaurant with the same visual characteristics that present contrary ratings for different users. By having so little data for each user and presenting a highly marked class imbalance, the interclass separation may not be sufficiently defined with this input data. And this type of situation is usually common in the field of online recommendations, where users do not usually offer a rich and sufficiently representative view of their tastes. For this reason, our model is better predicting user preferences when the architecture is deeper, because it needs to make a great computational effort to learn how to detect so many different cases with similar characteristics and very few examples.

As future work, we plan to test the adequacy of performing the feature extraction step depending on restaurants, rather than on users, taking as a reference the images of the restaurants and their overall rating. The idea is to learn to detect relevant characteristics using the general taste, and then apply this knowledge in a personalized way to each particular user. Our future research also involves the integration of the proposed model into a more complete RS that takes into account additional information, such as the text included in the reviews or sociodemographic data. Finally, we are also considering to apply our method to other RS that deal with images.

\bibliography{IEEEtran}
\end{document}